\documentclass[letterpaper]{article} % DO NOT CHANGE THIS
\usepackage{aaai2026}  % DO NOT CHANGE THIS
\usepackage{times}  % DO NOT CHANGE THIS
\usepackage{helvet}  % DO NOT CHANGE THIS
\usepackage{courier}  % DO NOT CHANGE THIS
\usepackage[hyphens]{url}  % DO NOT CHANGE THIS
\usepackage{graphicx} % DO NOT CHANGE THIS
\usepackage{natbib}  % DO NOT CHANGE THIS AND DO NOT ADD ANY OPTIONS TO IT
\usepackage{caption} % DO NOT CHANGE THIS AND DO NOT ADD ANY OPTIONS TO IT
\usepackage{algorithm}
\usepackage{booktabs}       % professional-quality tables
\usepackage{amsfonts}       % blackboard math symbols
\usepackage{nicefrac}       % compact symbols for 1/2, etc.
\usepackage{microtype}      % microtypography
\usepackage{xcolor}         % colors

\usepackage{rotating}
\usepackage{tcolorbox}
\usepackage{algorithm}
\usepackage{algpseudocode}
\usepackage{amsmath}
\usepackage{algorithmicx}
\usepackage{booktabs}
\usepackage{multirow}
\usepackage{amsfonts}
\usepackage{cleveref}
\usepackage{subcaption}
\usepackage{newfloat}
\usepackage{listings}
\DeclareCaptionStyle{ruled}{labelfont=normalfont,labelsep=colon,strut=off} % DO NOT CHANGE THIS
\floatstyle{ruled}
\newfloat{listing}{tb}{lst}{}
\floatname{listing}{Listing}
\title{Computer Vision Modeling of the Development of Geometric and Numerical Concepts in Humans}
\author{
    Zekun Wang,
    Sashank Varma
}
\affiliations{
    Georgia Institute of Technology\\
    School of Interactive Computing\\
    Atlanta, GA 30332 USA\\
    \{zekun,varma\}@gatech.edu
}

\usepackage{bibentry}
\begin{document}

\maketitle

\begin{abstract}
Mathematical thinking is a fundamental aspect of human cognition. Cognitive scientists have investigated the mechanisms that underlie our ability to thinking geometrically and numerically, to take two prominent examples, and developmental scientists have documented the trajectories of these abilities over the lifespan. Prior research has shown that computer vision (CV) models trained on the unrelated task of image classification nevertheless learn latent representations of geometric and numerical concepts similar to those of adults. Building on this demonstrated cognitive alignment, the current study investigates whether CV models also show developmental alignment: whether their performance improvements across training to match the developmental progressions observed in children. In a detailed case study of the ResNet-50 model, we show that this is the case. For the case of geometry and topology, we find developmental alignment for some classes of concepts (Euclidean Geometry, Geometrical Figures, Metric Properties, Topology) but not others (Chiral Figures, Geometric Transformations, Symmetrical Figures). For the case of number, we find developmental alignment in the emergence of a human-like ``mental number line'' representation with experience. These findings show the promise of computer vision models for understanding the development of mathematical understanding in humans. They point the way to future research exploring additional model architectures and building larger benchmarks.
\end{abstract}

\begin{links}
    % \link{Code}{https://aaai.org/example/code}
    % \link{Datasets}{https://aaai.org/example/datasets}
    % \link{Extended version}{https://aaai.org/example/extended-version}
\end{links}

\section{Introduction}

%%% SV: Can we get the citation/reference to be to the original Gelernter paper, which wa spublished in 1959? Here is the DOI for that paper: https://doi.org/10.1016/S0019-9958(59)90090-7
%%%
%%% deepthink2025: https://deepmind.google/discover/blog/advanced-version-of-gemini-with-deep-think-officially-achieves-gold-medal-standard-at-the-international-mathematical-olympiad/
Mathematical thinking is a fundamental aspect of human cognition, and as such has long been a target for AI researchers. Among the earliest AI programs were the Logic Theorist~\cite{1056797}, which proved theorems from \emph{Principia Mathematica}, and Gelernter’s geometry theorem prover~\cite{Gelernter1959}. There followed 60 years of steady progress on automating logico-mathematical reasoning, mostly within the symbolic paradigm. Over the past 10 years, rapid developments in ML have brought new successes to building systems that can reason mathematically. For example, in July 2025, the Gemini DeepThink model was able to meet the gold medal standard in the International Mathematical Olympiad~\cite{deepmindAdvancedVersion}. 

%%% Shah et al., 2023: Shah, R. S., Marupudi, V., Koenen, R., Bhardwaj, K., & Varma, S. (2023, July). Numeric magnitude comparison effects in large language models. In Findings of the Association for Computational Linguistics: ACL 2023 (pp. 6147–6161). Toronto, Canada.
%%% Stoianov \& Zorzi, 2012: Stoianov, I., & Zorzi, M. (2012). Emergence of a'visual number sense'in hierarchical generative models. Nature Neuroscience, 15, 194-196.
%%% Testolin et al., 2020: Testolin, A., Zou, W. Y., & McClelland, J. L. (2020). Numerosity discrimination in deep neural networks: Initial competence, developmental refinement and experience statistics. Developmental Science, 23, e12940.
AI is both an engineering discipline and a scientific discipline. As the field develops more and more performant systems, we must also ask whether these systems represent mathematical concepts in the same ways people do. If so, then these systems can be brought into cognitive science as models of human mathematical thinking. In fact, this is increasingly the case. There is a long history in cognitive science of studies of the mental representations and processes by which people reason mathematically. Research over the past decade has shown that computer vision (CV) models and LLMs represent geometric and numerical concepts similarly to people~\cite{shah-etal-2023-numeric,Stoianov2012,Testolin2020} 
% (Shah et al., 2023; Stoianov \& Zorzi, 2012; Testolin et al., 2020).

%%% Frank \& Goodman, 2025: Frank, M. C., & Goodman, N. D. (2025). Cognitive modeling using artificial intelligence. Annual Review of Psychology.
%%% Shah et al., 2024: Shah, R. S., Bhardwaj, K., & Varma, S. (2024, November). Development of cognitive intelligence in pre-trained language models. Proceedings of the 2024 Empirical Methods in Natural Language Processing Conference (pp. 9632–9657). Miami, Florida: Association for Computational Linguistics.
%%% Warstadt \& Bowman, 2022: Warstadt, A., & Bowman, S. R. (2022). What artificial neural networks can tell us about human language acquisition. In S. Lappin & J.-P. Bernardy (Eds.) Algebraic Structures in Natural Language (pp. 17-60). CRC Press.
%%% K. He et al., 2016: He, K., Zhang, X., Ren, S., & Sun, J. (20016) Deep residual learning for image recognition. In Proceedings of the IEEE Conference on Computer Vision and Pattern Recognition, 770–778.
The vast majority of these studies have investigated the cognitive alignment between ML models and adult thinking. Here, we evaluate their potential developmental alignment: Does their improving mathematical performance across training match the developmental progressions observed in children? Researchers are only just beginning to move beyond the question of cognitive alignment to the question of developmental alignment~\cite{Frank2025,shah-etal-2024-development,warstadt2024artificialneuralnetworkstell}. In this case study, we train a ResNet-50 model~\cite{he2015deepresiduallearningimage} on the ImageNet image dataset~\cite{imagenet}, measure as its sensitivity to geometric concepts grows and the precision of its number representations sharpens across checkpoints, and compare these progressions to those observed in children and adults across the lifespan.

\subsection{Literature Review}

The current study focuses on geometric and topological (GT) concepts and on number representations. This section reviews cognitive science studies of how adults understand geometry and number, and developmental science studies of the trajectories by which they come to this understanding. It also reviews investigations of whether ML models can capture these cognitive and developmental patterns. Although some of this work has been done with LLMs, we focus on CV models because this is the class of models explored in the current study.

\paragraph{Geometric and Topological Concepts}

\begin{figure}
    \centering
    \includegraphics[width=0.85\linewidth]{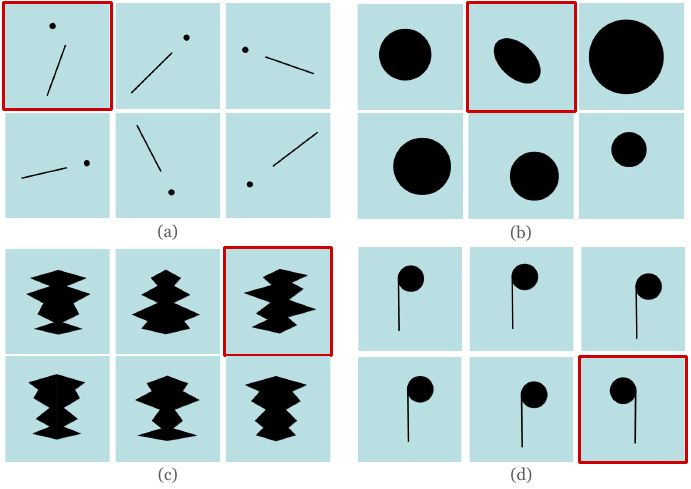}
    \caption{Sample stimuli for 4 GT concepts: (a) Euclidean Geometry - alignment of points in lines. (b) Geometrical Figures - circle. (c) Symmetrical Figures - vertical axis. (d) Chiral Figures - vertical axis. The odd-one-out is indicated by the red box.}
    \label{fig:gt-stimuli}
\end{figure}

%%% SV (NEW): The better citation for footnote 1 is Marupudi et al. (2023), not Upadhyay et al. (2025). The Marupudi paper is not currently cited in the paper. Can you add this citation and reference? (If doing so puts us over the page limit, then ignore this request.) Done - Zekun.
%%%
%%% Marupudi \& Varma, 2023: Marupudi, V., & Varma, S. (2023). Graded human sensitivity to geometric and topological concepts. Cognition, 232, e105331.
%%%
%%%
%%% SV: We can eliminate the footnotes to save space if needed for the Acknowledgments section.
%%%
%%% SV: You use "\citet" for in-line citations. Is this the correct style for ACM? I am used to using "\citet". Please confirm which one AAAI wants.
%%%
%%% Dehaene et al. (2006): Dehaene, S., Izard, V., Pica, P., & Spelke, E. S. (2006). Core knowledge of geometry in an Amazonian indigene group. Science, 311, 381-384.
%%% Spelke \& Kinzler, 2007: Spelke, E. S., & Kinzler, K. D. (2007). Core knowledge. Developmental Science, 10, 89–96.
%%% Shepard, 2001: Shepard, R. N. (2001). Perceptual-cognitive universals as reflections of the world. Behavioral and Brain Sciences, 24, 581-601.
The seminal study of how humans understand GT concepts is by~\citet{Dehaene2006}. They developed an odd-one-out task that tests people’s sensitivity to 43 concepts, which themselves group into 7 classes: Topology, Euclidean Geometry, Geometric Figures, Symmetrical Figures, Chiral Figures, Metric Properties, and Geometrical Transformations. Figure~\ref{fig:gt-stimuli} shows example stimuli for concepts from 4 classes. For each stimulus, the task is to judge which of the 6 images is the “odd one out”. The images differ on multiple perceptual dimensions. Critically, 5 of the images embody the target concept whereas the 6th one does not. If people are sensitive to that concept, they will be above chance ($1/6$) in selecting that image as the odd one out. \citet{Dehaene2006} administered this task to adults and children from the Mundurucu, an Amazon river valley group whose members have little or no formal schooling and are therefore unlikely to have received explicit instruction on these GT concepts. Nevertheless, they were above chance in selecting the odd-one-out for 37 of the 43 concepts (86\%). The researchers also tested Western participants, finding that the children performed as well as the Mundurucu adults and children, and that the adults performed slightly better.\footnote{The findings with Western adults have been replicated (e.g., \citet{Marupudi2023}).} They interpreted the strong performance of the Mundurucu as evidence that people have \emph{core knowledge} of GT concepts, which is to say they are part of the human endowment~\cite{Spelke2007}.\footnote{It would make sense for evolution to deliver such an endowment given that the universe, and more locally the terrestrial environment, is governed by geometry and topology~\cite{Shepard2001}.} 

%%% SV: This footnote is perhaps too cute. Eliminate it first if necessary to save space for the Acknowledgments section.
%%%
%%% Greenough et al., 1987: Greenough, W. T., Black, J. E., & Wallace, C. S. (1987). Experience and brain development. Child Development, 58, 539-559.
%%% Izard and Spelke (2009): Izard, V., & Spelke, E. S. (2009). Development of sensitivity to geometry in visual forms. Human Evolution, 23, 213-248.
Even from a strong core knowledge position, not all GT concepts need be part of a child’s initial repertoire.\footnote{We do not spring fully-formed from the brow of Zeus, like the goddess Athena.} Rather, it is possible that some are available very early whereas others appear later, perhaps because they are learned through experience~\cite{Greenough1987}. \citet{Izard2009-em} documented the developmental progression of GT concepts. In their Experiment 1, Western children ages 3-6 years old completed the odd-one-out task. The children showed above-chance sensitivity to 27 (63\%) of the 43 concepts, suggesting that while some concepts might be part of core knowledge and available very early on, other concepts might be learned from experience in the world (including formal mathematics instruction). For example, the young children showed sensitivity to all 8 of the Euclidean Geometry concepts -- but to none of the 8 Geometric Transformations concepts.

%%% SV: We can eliminate the footnote to save space if needed for the Acknowledgments section.
%%%
Recently, researchers have asked whether CV models are sensitive to GT concepts~\cite{hsu2022geoclideanfewshotgeneralizationeuclidean,campbell2024humanlikegeometricabstractionlarge}. This is an interesting question because CV models are not trained to learn about mathematics. Rather, they are trained to accurately classify images. Thus, they can be understood as instantiating the view that (perceptual) development is mostly a matter of learning, which contrasts with the strict core knowledge view~\cite{Spelke2007}. This raises the question of whether, as a ``side effect'' of learning to classify images, CV models also become sensitive to GT concepts? If so, then the view of development as learning may be largely sufficient, and there may be less need to posit a role for core knowledge.\footnote{Of course, we can still ask whether the training of CV models and the (perceptual) development of children are analogous. We return to this question in the General Discussion.} 

%%% Upadhyay et al. (2025): Upadhyay, N., Marupudi, V., Varma, K., & Varma, S. (2025, February). Alignment of CNN and human judgments of geometric and topological concepts. In Proceedings of the Thirty-Ninth AAAI Conference om Artificial Intelligence (AAAI-25) (pp. 1556-1564), Philadelphia, PA.
%%% Wang and Varma (2025): Wang, Z., & Varma, S. (2025, July). Computer vision models show human-like sensitivity to geometric and topological concepts. In Proceedings of the 47th Annual Conference of the Cognitive Science Society (pp. XXX-XXX). San Francisco, CA.
\citet{Upadhyay2025} tested 5 CNN models on the odd-one-out task. The best performing model, ResNet-18 \cite{he2015deepresiduallearningimage}, showed sensitivity to 17 (40\%) of the 43 GT concepts. This absolute level of performance was disappointing: though above chance (again, $1/6$), it was below that of the young children tested by \citet{Izard2009-em}, who recall were sensitive to 27 (63\%) of the 43 GT concepts. More promising was the correlation between the performance of the model and of the children at the level of the 7 classes of GT concepts, which was medium in size ($r = 0.52$, $p > .20$).
%%% or p<0.2?
This suggests that the model and the young children found the same classes of concepts relatively easy vs. difficult. \citet{wang2025computervisionmodelshumanlike} replicated these findings and extended them beyond CNNs to other model architectures: vision transformers and vision-language models. The vision transformer models they tested, ViT and DINOv2, achieved overall accuracies ($47\%$ and $49\%$, respectively) closer to the young children tested by Izard and Spelke (2009). Moreover, the correlations between the models and the young children across the 7 classes were exceptionally high: $r=0.93$ and $r=0.91$ ($p$s $< 0.01$), respectively.

\paragraph{Number Representation}

\begin{figure}
    \centering
    \includegraphics[width=1.01\linewidth]{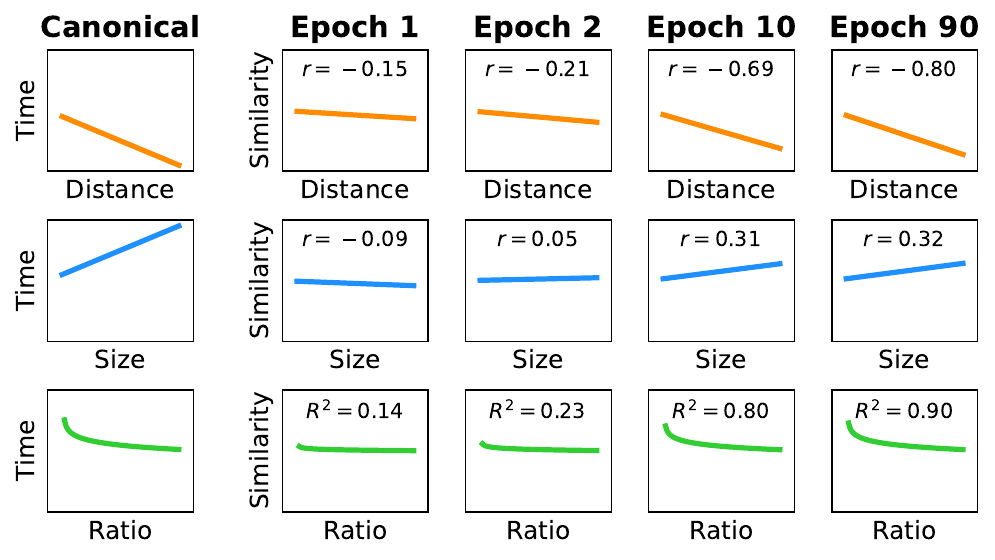}
    \caption{Idealized distance, size, and ratio effects observed in humans (left) and the emergence of these effects in ResNet-50 over training (right).}
    \label{fig:num-stimuli}
\end{figure}

%%% SV: The Moyer and Landauer (1967) citation and reference are rendering in ALL CAPS.
%%%
%%% Whalen, 1999: Whalen, J., Gallistel, C. R., & Gelman, R. (1999). Nonverbal counting in humans: The psychophysics of number representation. Psychological Science, 10, 130–137.
%%% Moyer \& Landauer, 1967: Moyer, R. S., & Landauer, T. K. (1967). Time required for judgments of numerical inequality. Nature, 215, 1519–1520.
%%% Parkman, 1971: Parkman, J. M. (1971). Temporal aspects of digit and letter inequality judgments. Journal of Experimental Psychology, 91, 191–205.
%%% Gallistel & Gelman, 1992: Gallistel, C. R., & Gelman, R. (1992). Preverbal and verbal counting and computation. Cognition, 44, 43–74.
%%% Piazza et al., 2007: Piazza, M., Pinel, P., Le Bihan, D., & Dehaene, S. (2007). A magnitude code common to numerosities and number symbols in human intraparietal cortex. Neuron, 53, 293–305.
%%% Halberda et al., 2008: Halberda, J., Mazzocco, M. M. M., & Feigenson, L. (2008). Individual differences in non-verbal number acuity correlate with maths achievement. Nature, 455, 665–668.
%%% SM1977 CITE: Sekuler, R., & Mierkiewicz, D. (1977). Children’s judgments of numerical inequality. Child Development, 48, 630–633.
%%% Moore \& Ashcraft, 2015: Moore, A. M., & Ashcraft, M. H. (2015). Children’s mathematical performance: Five cognitive tasks across five grades. Journal of experimental child psychology, 135, 1-24.
Cognitive science research has shown that people understand numbers by reference to a mental number line that is psychophysically scaled~\cite{Whalen1999}. Evidence for this representation comes from three effects, depicted in idealized form in the left panel of Figure~\ref{fig:num-stimuli}. The \emph{distance effect} is that when comparing which of two numbers $n_1$ and $n_2$ is greater, judgment time decreases linearly with the distance $|n_1 - n_2|$ between them~\cite{MOYER1967}. This is consistent with people locating the two numbers on their MNL and then discriminating which is one is ``to the right'' of the other. The farther apart they are, the easier the discrimination. The \emph{size effect} is that when comparing two numbers, the greater their average size $(n_1 + n_2)/ 2$, the slower the judgment time~\cite{Parkman1971}. For example, people take longer to compare 8 vs. 9 than 1 vs. 2 even though the distance is the same in both cases. This suggests that the distance between numbers is not constant, as in the conventional number line of mathematics, but psychophysically compressed, decreasing as numbers get larger. These two effects are combined in the ratio effect, which is that the time to compare two numbers decreases as the ratio of the larger to the smaller (i.e., $\max(n_1, n_2)/\min(n_1, n_2)$) increases~\cite{Gallistel1992}. For example, people are very slow to compare 8 vs. 9 (ratio $= 1.125$), a bit faster to compare 1 vs. 2 (ratio $= 2.0$), and faster still to compare 1 vs. 9 (ratio $= 9.0$). People show these effects whether numbers are presented as digits (e.g., `3'), words (e.g., `three'), or numerosities (e.g., `$\circ \circ \circ$')~\cite{Piazza2007}. Importantly, the precision of the MNL improves over development, which can be seen in the sharpening of the distance, size, and ratio effects as children get older~\cite{Halberda2008-qa,Sekuler1977,Moore2015-yq}.

%%% Brannon \& Terrace, 1998: Brannon, E. M., & Terrace, H. S. (1998). Ordering of the numerosities 1 to 9 by monkeys. Science, 282, 746-749. 
%%% Nieder, 2021: Nieder, A. (2021). The evolutionary history of brains for numbers. Trends in Cognitive Sciences, 25, 608-621.
%%% Zorzi \& Stoianov, 2018: Zorzi, M., & Testolin, A. (2018). An emergentist perspective on the origin of number sense. Philosophical Transactions of the Royal Society B: Biological Sciences, 373(1740), 20170043.
%%% Nasr et al., 2019: Nasr, K., Viswanathan, P., & Nieder, A. (2019). Number detectors spontaneously emerge in a deep neural network designed for visual object recognition. Science Advances, 5, eaav7903.
The presence of these effects very early in development, and even in other species, has led to the proposal that the MNL is ``evolutionarily ancient''~\cite{Brannon1998-ml,Nieder2021-iw}. Alternatively, this representation might not be part of the human endowment, but rather learned ``for free'' through experience in the visual world. CV models can be used to test the sufficiency of this learning account. \citet{Stoianov2012}, in an early modeling study, trained a deep neural network on images depicting numerosities. The representations the model learned showed the distance and ratio effects, consistent with the model having learned a latent MNL. Subsequent studies with deep neural networks showed that this representation sharpens over training, paralleling its developmental trajectory in humans~\cite{Testolin2020,Zorzi2017-wx}. More recent research utilizing conventional CV models -- CNNs trained on ImageNet -- has also found evidence of a latent MNL representation~\cite{Nasr2019-xl}.

%%% Upadhyay & Varma (2023): Upadhyay, N., & Varma, S. (2023, December). CNN models’ sensitivity to numerosity concepts. Poster presented at the 3rd Workshop on Mathematical Reasoning and AI (MATH-AI 23), NeurIPS’23, New Orleans, LA.
%%% Simonyan \& Zisserman, 2015: Simonyan, K., & Zisserman, A. (2015) Very deep convolutional networks for large-scale image recognition. In International Conference on Learning Representations.
In the closest prior study, \citet{upadhyay2023cnn} evaluated the latent number representations of multiple pretrained CNNs such as VGG19~\cite{simonyan2015deepconvolutionalnetworkslargescale}. They presented images of the numerosities $1-9$ and read off the vector representation on the final fully-connected layer of these models. VGG19 showed strong distance, size, and ratio effects, signaling that an MNL representation had been learned. They used multidimensional scaling (MDS) to reconstruct this representation, finding that it differed from the canonical MNL only in switching the positions of 1 and 2.

\subsection{Research Questions}

Previous research has established the sensitivity of CV models to GT concepts and has shown that CNN models possess latent number representations similar to the MNL of humans. With one notable exception (e.g., \citet{Testolin2020}), this research has focused on the question of cognitive alignment, i.e., the correspondence of models to adult thinking. Here, we ask the question of developmental alignment:
\begin{enumerate}
    \item Over training, does the sensitivity of ResNet-50 to GT concepts increase, and does this increase follow the developmental trajectory observed in people?
    \item Over training, do the number representations of ResNet-50 increasingly show the distance, size, and ratio effect that signal an MNL representation, and does the precision of this representation improve according to the developmental trajectory observed in children?
\end{enumerate}

\section{Experiment 1}

Experiment 1 investigated research question (1).

\subsection{Method}

\paragraph{Model and Training}
%%% Kriegeskorte, 2015: Kriegeskorte, N. (2015). Deep neural networks: a new framework for modeling biological vision and brain information processing. Annual review of vision science, 1, 417-446.
%%% Yamins \& DiCarlo, 2016: Yamins, D. L., & DiCarlo, J. J. (2016). Using goal-driven deep learning models to understand sensory cortex. Nature neuroscience, 19, 356-365.
For this case study, we chose the ResNet-50 model \cite{he2015deepresiduallearningimage} because \citet{Upadhyay2025} found that among the 5 CNNs they tested, it showed the greatest sensitivity to GT concepts and also moderate alignment with young children.\footnote{The subsequent study by \citet{wang2025computervisionmodelshumanlike} found that the vision transformer models ViT~\cite{dosovitskiy2021imageworth16x16words} and DINOv2~\cite{oquab2024dinov2learningrobustvisual} showed better overall performance and stronger developmental alignment than CNNs. However,  we were unable to locate training checkpoints for either of these models and lacked the compute budget to train them ourselves.} Furthermore, the architecture of CNNs maps closely to that of the human visual system, making them better candidates as cognitive (neuro)science models than other CV model architectures~\cite{Kriegeskorte2015,Yamins2016-ib}.

In greater detail, our network followed the standard ResNet-50 configuration: a $7{\times}7$ conv (64 channels, stride~2) + BN/ReLU, $3{\times}3$ max-pool (stride~2), four residual stages with bottleneck blocks in the pattern $[3,4,6,3]$ and output widths $[256,512,1024,2048]$, global average pooling, and a 1,000-way fully connected classifier (about 25.6M parameters). We trained on ImageNet-1k (ILSVRC-2012) \cite{imagenet} using the official train/validation split (1.28M/50k images). 
Training images were processed following the original implementation with \texttt{RandomResizedCrop} to $224{\times}224$ (scale $[0.08,1.0]$, aspect ratio $[\tfrac{3}{4},\tfrac{4}{3}]$), random horizontal flip ($p{=}0.5$), and per-channel normalization to ImageNet mean/std.
Validation resized images to $256\times 256$ and then $224{\times}224$ center-cropped with identical normalization. 
We optimized cross-entropy loss between the predicted class label and the actual labels with SGD, training for 90 epochs with global batch size 256, at an initial learning rate of $0.1$
A step-scheduler was used to decrease learning rate by a factor of $0.1$ every 30 epochs, ending training at a learning rate of $1\times10^3$.
Runs use PyTorch on a single A40 (48 GB) GPU. We saved a full checkpoint (weights, optimizer/scheduler state, RNG) at the end of every epoch.
Developmental analyses below use the sequence of checkpoints at saved epochs. 
Validation accuracy after training matches the standard ResNet-50 reference (top-1 $\sim$76\%, top-5 $\sim$93\%), confirming that our model is comparable to widely reported baselines and suitable for subsequent developmental alignment evaluations.

\paragraph{Design and Materials}
%%% Table XXX of the Supplementary Materials: must be created. I think there is a model for this in the supplementary materials for Marupudi and Varma (2023) or else Upadhyay et al. (2025).
The stimuli were from \citet{Dehaene2006}. As described above, there is one stimulus for each of 43 GT concepts (e.g., ‘holes’); see Figure~\ref{fig:gt-stimuli} for examples.\footnote{We thank Dr. Stanislas Dehaene for providing the stimulus images from this study. For further information about this dataset, please contact him directly.}  Each stimulus is composed of 6 images where 5 embody the GT concept and 1 does not. The task is to choose the ‘odd one out’. The correct choice is the image that does \emph{not} embody the GT concept, and so chance is $1/6$. The 43 GT concepts can be aggregated into 7 broader classes: Topology, Euclidean Geometry, Geometrical Figures, Symmetrical Figures, Chiral Figures, Metric Properties, and Geometrical Transformations. See Table 1 of the Supplementary Materials for a listing of all GT concepts and the classes to which they belong.

\paragraph{Human Data}
%%% Supplementary Materials Figure XXX: You have a draft of this in the presentation slides you shared with me.
%%% Figure XXX of the Supplmentary Materials: You have a draft of this in the presentation slides you shared with me. Remember to reformat the 2x4 so that the top-left panel contains the overall accuracy data and the remaining panels the data for each of the 7 classes.
The human data were from Experiment 2 of \citet{Izard2009-em}, which investigated the development of sensitivity to GT concepts across the lifespan. That study tested 400 Western participants ages $6-51$ years old. Most of the participants were children, adolescents, or young adults (i.e., 28 years old or younger); see the Supplementary Materials Figure 1 for a histogram of participant ages. Participants completed 2 practice trials followed by 43 experimental trials. On each trial, a stimulus was shown and participants clicked their choice of the odd-one-out image. 
%% I think Figure 3 already shown these results. Maybe we can remove the following sentence or refer to Figure 3? -Zekun
% The data were then aggregated: the average overall accuracy was computed for each age, as were the average accuracies for each of the 7 classes. See Figure XXX of the Supplmentary Materials for graphs of these data. 

\paragraph{Procedure}
After each training epoch, we ran the model on the odd-one-out task, following the same method of \citet{Upadhyay2025} and \citet{wang2025computervisionmodelshumanlike}. For each stimulus, each of the 6 images was first rescaled and cropped to 224×224 pixels. Each image was passed through the model and the representation before the final prediction layer collected as a vector of 2048 activations. Next, the cosine similarity between each pair of image vectors was computed. The model’s choice of the odd-one-out image was the one with the lowest
average cosine similarity to the other 5 images. We aggregated the model’s performance to compute its overall accuracy and its accuracy for each of the 7 classes. These measures exactly parallel those computed for the human data.

\subsection{Results}

In the visualizations and analyses that follow, we mapped 2 epochs of model training to 1 year of human development. This was the natural mapping as the model was trained for 90 epochs and the age range of the sample was 45 years.

%%% SV: The statistic you report for the first "correlation" is not "r" but "R^2". We should report "r". I see two reasons for this discrepancy. One is that the statistics reported is, in fact, R^2. If this is the case, the fix is easy, since r = sqrt(R^2). So r would be .71. The other is that you mistyped "r" as "R^2". If this is the case, then simply fix this error.
%%%
% Ah! That's Pearson's r. It is a typo. I double checked my code and I was doing Pearson's r. -Zekun

%%% SV: The text desribing how the correlation is computed. Ism't this the old text? Shouldn't it be updated to: "the correlation between them over years $6 – 51$ corresponding to epochs $2, 4, ..., 90$"
% fixed - Zekun

\begin{figure}
    \centering
    \includegraphics[width=0.75\linewidth]{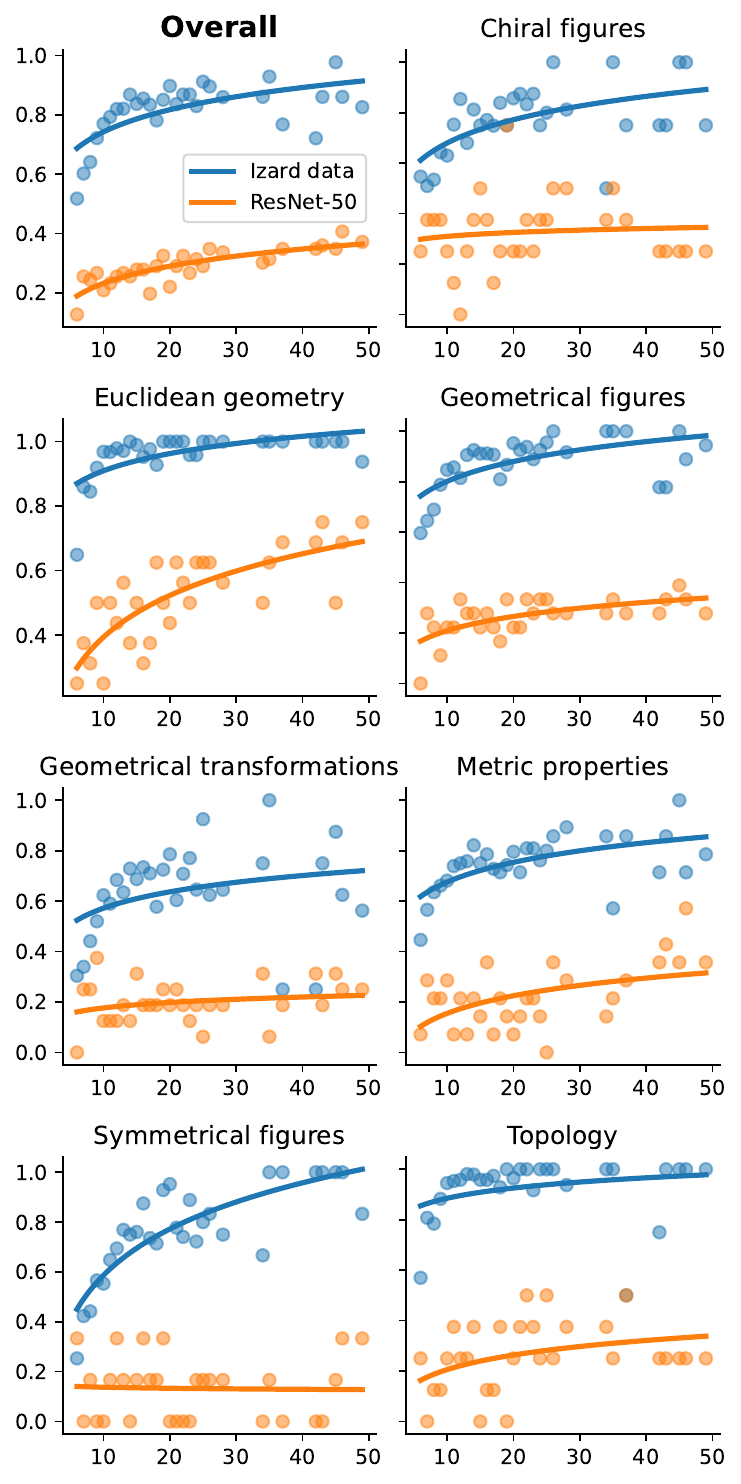}
    \caption{Overall accuracy function for humans over development and for ResNet-50 over training (top-left panel), and for these data broken down for each of the 7 classes of GT concepts (remaining panels). $x$-axis: age (epochs); $y$-axis: accuracy.}
    \label{fig:gt-dev-results}
\end{figure}

%%%
%%% SV: I think we want to redo the stats, so that you are relating years 6, 7, 8, ..., 51 to epoichs 2, 4, 6, ..., 90.
%%% -Yep, already did. I averaged every two epochs. Like moving average of window size=2, stride=2. -Zekun
%%% SV: Nice!
%%%
%%% SV: We may be able to cut the last sentence/analysis, which is about the RMSD between the two lines. Visual inspection is probably enough evidence for the statement.
%%% Sounds good! -Zekun
%%%
%%% Logan, 1988: Logan, G. D. (1988). Toward an instance theory of automatization. Psychological review, 95(4), 492-527.
%%% Laird et al., 1986: Laird, J. E., Rosenbloom, P. S., & Newell, A. (1986). Chunking in Soar: The anatomy of a general learning mechanism. Machine Learning, 1, 11-46. 
The top-left panel of Figure~\ref{fig:gt-dev-results} shows the overall accuracy curves for humans over development and for ResNet-50 over training. Both humans and the model show improving performance with experience: the Pearson's correlation between them over years $6-51$ corresponding to epochs $2, 4, 6, \cdots, 90$ is $r=0.50$ ($p<0.01$). Because many human learning curves follow a power function~\cite{Laird1986,Logan1988}, we fit a power function to each set of data. This function offered a good account of the human data ($R^2=0.40$) and also of the model data ($R^2 = 0.66$), giving further evidence of developmental alignment. That said, humans decisively outperform the model.
%; the RMSD between the two, again over the corresponding years / epochs, is XXX.

%%% SV: The new Figure 3 and the new Table 2 in the Supplementary Materials both look great!
%%%
%%% Table XXX of the Supplementary Materials: You'll have to create this and add it to the supplementart materials. I think adding the correlation to each of 7 panels and the R^2 to each of the 7 x 2 power function fits to the panels might be too visually crowded. If you have the time, you can try, but don't prioritize this I think.
%%% SV: As with the previous paragraph, we will likely dispense with the RMSDs and just rely on visual inspection by the reader.
The remaining panels of Figure~\ref{fig:gt-dev-results} show the average accuracy curves over development / training for each of the 7 classes. Humans show improving performance across development for all 7 classes, with the rate of improvement following a power function. By contrast, the model shows improving performance for only 4 of the 7 classes: Euclidean Geometry, Geometric Figures, Metric Properties, and Topology. (And for these 4 classes, model performance again lags human performance.) For the remaining 3 classes -- Chiral Figures, Geometric Transformations, and Symmetrical Figures -- the model’s accuracy is both low and hardly improves across training. The associated correlations between human and model performance for each of the 7 classes as well as the fits of power functions can be found in Table 2 of the Supplementary Materials.

\subsection{Discussion}

Experiment 1 investigated research question (1): Whether ResNet-50’s growing sensitivity to GT concepts over training matches the trajectories observed in humans over development. The model’s overall performance on the odd-one-out task improves with training according to a power function, matching the trajectory observed in humans – although the model’s absolute level of performance is lower. At a finer-grain level, the model shows growing sensitivity for 4 of the 7 classes. This suggests that Euclidean Geometry, Geometric Figures, Metric Properties, and Topology concepts might come ``for free'' when learning to perceive the visual world, and need not be entirely located within core knowledge. By contrast, for Chiral Figures, Geometric Transformations, and Symmetrical Figures concepts, the model shows almost no improvement over training. This stands in contrast to humans, who show improved sensitivity over development. This is evidence that these concepts do not come ``for free'', and instead might be part of core knowledge or be learned through explicit mathematics instruction.

\section{Experiment 2}

Experiment 2 investigated research question (2).

\subsection{Method}

\paragraph{Model and Training}
Same as Experiment 1.

\paragraph{Design and Materials}

\begin{figure}
    \centering
    \includegraphics[width=1\linewidth]{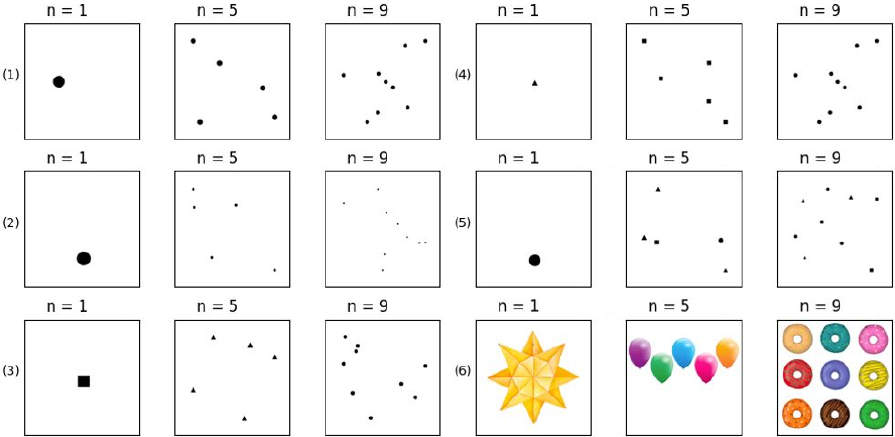}
    \caption{Example stimuli for the stimulus sets $1-6$.}
    \label{fig:num-sti}
\end{figure}

%%% SV: This is a small request, but it would nice to state the pixel dimensions of the stimuli, of the bounding box. This is relevant context for when we explain how many of the pixels are black in stimulus set 1.
%%%
%%% Zekun please fix the latex math formatting for stimulus set (4).
%%% Figure 5: Need to create a new/nice figure showing sample stimuli.
The stimuli were from \citet{upadhyay2023cnn}. Each is a $720\times720$-pixel image showing a numerosity of $1-9$ items. The stimuli are organized into 6 sets that vary in which perceptual variables are controlled, which are varied parametrically, and which are allowed to vary randomly. The stimulus sets are intended to be progressively more difficult for models, to enable titration of their sensitivity to numerosity (over perceptual variables).
\begin{enumerate}
    \item The items are black circles randomly placed on a white background. For a given area $A$, the total area of each numerosity (i.e., the number of black pixels) is controlled to be $A$. Thus, a stimulus with numerosity 1 and another with numerosity 9 each have A black pixels. This prohibits using this perceptual feature (total area) as a proxy for numerosity. The total area is parametrically varied across five levels $A_1$ – $A_5$ corresponding to 103 – 518 black pixels, defining five subsets of images.
    \item Like (1) but the total circumference $C$ is controlled, so that this feature cannot be used as a proxy for numerosity. The total circumference is parametrically varied across five levels $C_1$ – $C_5$ corresponding to 100 – 300 pixels, defining five subsets of images.
    \item Like (2) but the items of the two numerosities are randomly varied, e.g., circles in one image and squares in another. This enables testing generalization across shapes.
    \item Like (3) but the total areas of the two numerosities is randomly varied, i.e., one having area $A_i$ and the other area $A_j$ ($i \neq j$). This enables generalization across both shapes and areas.
    \item ‘Anything goes’: Like (4) except the individual items of each numerosity are randomly varied so that each is a mixture of circles, squares and triangles of different areas. This enables further generalization of the findings.
    \item Naturally occurring numerosities found using Google Search (and manually verified). These are mostly stylized like clip art. These items vary on many perceptual dimensions (e.g., shape, size, drawing style, color, etc.), enabling further generalization of the findings.
\end{enumerate}
See Figure \ref{fig:num-sti} for example stimuli from each set, and the Supplementary Materials for further details on their construction.

\paragraph{Procedure}
%%% Zekun please fix the latex math formatting.
After each training epoch, we evaluated the model’s distance, size, and ratio effects for each stimulus set. Recall these effects are collectively evidence for an MNL representation. Given that the numerosities are in the range $1-9$, there are $(9 \times 8) / 2 = 36$ pairs of numerosities $n_1$ and $n_2$ such that $n_1 \neq n_2$ (i.e., so that one is more numerous than the other). 

%%% Zekun please fix the latex math formatting.
%%% XXX layer: Please specify.
For each stimulus set (and for each level of total area $A$ or total circumference $C$, if relevant), for each of the 36 comparisons, we randomly sampled stimuli of numerosity $n_1$ and $n_2$. We passed each through the model and captured the vector representation before the final prediction layer. We then computed the cosine similarity between the two vectors. We made the following linking hypothesis to map model performance to human performance: the less similar the vectors, the more discriminable the corresponding numerosities, and thus the faster the predicted time to judge which one is the greater numerosity. This is the same linking hypothesis that has been used in prior studies of numerical alignment between humans and CV models \cite{upadhyay2023cnn} and LLMs \cite{shah-etal-2023-numeric}. The three effects were computed as follows:
\begin{itemize}
    \item distance effect: The correlation between the similarity of the vectors and the distance between the numerosities $|n_1-n_2|$. A negative correlation indicates a human-like distance effect.
    \item size effect: The correlation between the similarity of the vectors and the average size of the numerosities $(n_1 + n_2)/ 2$. A positive correlation indicates a human-like size effect.
    \item ratio effect: The $R^2$ of fitting a negative exponential function predicting the similarity of the vectors by the ratio of the larger numerosity to the smaller: $\max(n_1, n_2)/\min(n_1, n_2)$. A value closer to 1 indicates a human-like ratio effect.
Canonical distance, size, and ratio effects are shown in the left panel of Figure~\ref{fig:num-stimuli}. 
\end{itemize}

\subsection{Results}

Research question (2) asks if the number representations of ResNet-50 develop over training along the same trajectory as the MNL of humans sharpens over development. To visualize what this would mean, the right panel of Figure~\ref{fig:num-stimuli} plots the distance, size, and ratio effects after epochs 1, 2, 10, and 90 of training. (We chose these epochs because the model rapidly learns at the earlier training stages.) We see that the effects are absent early in training, signaling the absence of an MNL representation. However, over training, these effects manifest. Thus, as a ``side effect'' of learning to classify images, the model learns a human-like representation of number.

\begin{figure}[!ht]
    \centering
    \includegraphics[width=1\linewidth]{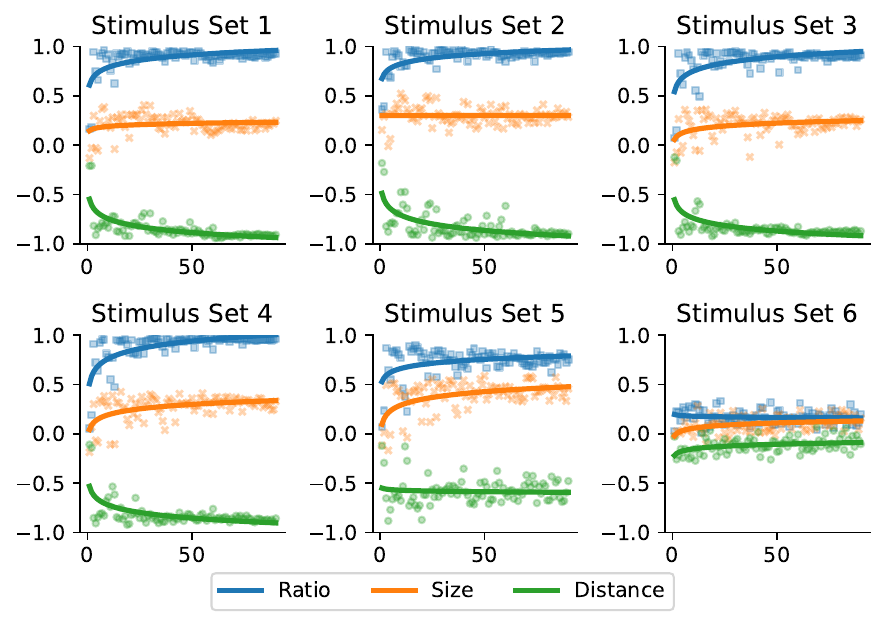}
    \caption{The distance effect correlation (green; expected to be close to $-1$), size effect correlation (orange; expected to be close to 1), and ratio effect $R^2$ (blue; fit of the negative exponential function, expected to be close to 1) of ResNet-50 over training.
    % The panel legends give the fit of a power function to the growth of each effect over training.
    $x$-axis: epochs; $y$-axis: correlation or $R^2$.}
    \label{fig:num-dev-results}
\end{figure}

%%% Zekun please fix the latex math formatting.
At a more detailed level, we can plot the trajectory of these effects over all 90 epochs. This is shown in Figure~\ref{fig:num-dev-results} -- the correlations for the distance and size effects and the $R^2$ for the ratio effect. We see that the distance effect is robust: it appears early in training, follows the canonical functional form (i.e., a negative correlation), and holds for all but the most varied stimulus sets (1 and 6). The ratio effect is also robust, following the canonical functional form (i.e., the $R^2$ is high) for all but the most varied stimulus sets (5 and 6). By contrast, the size effect is smaller in size, with correlations positive (as predicted) but closer to 0 than 1. Curiously, the size effect is weakest in the ‘easiest’ stimulus sets: 1 (equal-area circles) and 2 (equal-circumference circles).

%%% SV: The panel legends have disappeared, and so have the "the fit of a power function to the growth of each effect over training". I have edited the caption here accordingly.
%%%
%%% Zekun please fix the latex math formatting in the caption.

We conducted a growth curve analysis of the developmental trajectories in Figure~\ref{fig:num-dev-results}. Specifically, for each stimulus set, we fit a power function to each of the three effects. We refer the reader to Table 3 of the Supplementary Materials for the fits of power functions. The overall pattern is for developmentally plausible growth of the distance and ratio effects, with a power function characterizing the improvement of the (negative) correlation and the $R^2$ fit value, respectively, over training epochs. This holds for all but the most varied stimulus set (6). By contrast, the growth of the size effect is less human-like, with the power function offering a generally worse account of the improvement of the (positive) correlation over training epochs.

\begin{figure*}[!ht]
    \centering
    \begin{subfigure}[t]{.23\textwidth}
        \centering
        \includegraphics[width=1\linewidth]{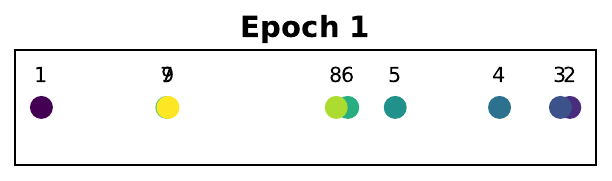}
        % \vspace*{-20pt}
        \caption{}
    \end{subfigure}
    ~
    \begin{subfigure}[t]{.23\textwidth}
        \centering
        \includegraphics[width=1\linewidth]{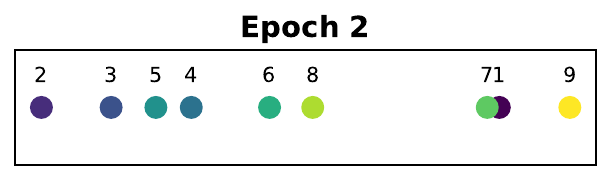}
        % \vspace*{-20pt}
        \caption{}
    \end{subfigure}
    ~
    \begin{subfigure}[t]{.23\textwidth}
        \centering
        \includegraphics[width=1\linewidth]{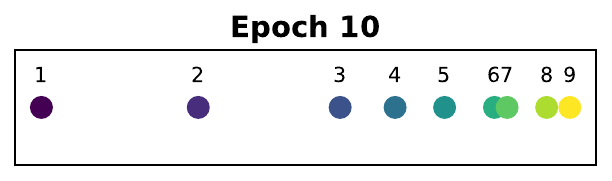}
        % \vspace*{-20pt}
        \caption{}
    \end{subfigure}
    ~
    \begin{subfigure}[t]{.23\textwidth}
        \centering
        \includegraphics[width=1\linewidth]{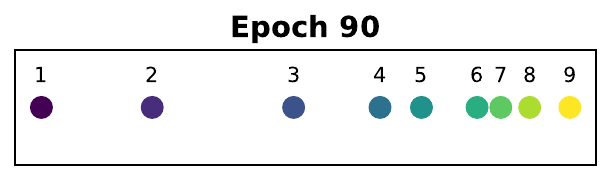}
        % \vspace*{-20pt}
        \caption{}
    \end{subfigure}
    \caption{Reconstructed number line representations of ResNet-50 over training showing the sharpening of its MNL.}
    \label{fig:mil}
\end{figure*}

%%% XXX: Specify the layer of ResNet-50.
Finally, we followed \citet{upadhyay2023cnn} and reconstructed the latent number line representation of the model at each epoch. Specifically, for stiumulus set (1), we formed a $9 \times 9$ matrix where each entry  is the cosine similarity between the vector representations of the corresponding numerosities before the final prediction layer. We submitted these pairwise similarities to MDS and requested a 1D solution, which we interpret as the model’s latent number line representation at that point in training. Figure~\ref{fig:mil} plots these for epochs 1, 2, 10, and 90. (Again, we chose these epochs because the model learns rapidly.) Over training, this representation comes to resemble the canonical MNL of humans, further showing the model’s developmental alignment.

\subsection{Discussion}

Experiment 2 investigates research question (2): Whether, over training, the number representations of ResNet-50 increasingly show the distance, size, and ratio effects that signal an MNL representation? This was the case. Across a range of stimulus formats, the model showed the distance and ratio effects early in training, and these effects only strengthened over time. There was less evidence for the orderly emergence of the size effect, and its generalization was lower to more varied stimulus presentation formats. Finally, a reconstruction of the model's number line representation over training shows the increasing sharpening of its MNL, further evidencing its developmental alignment. These findings support the proposal that visual experience in the world may deliver an MNL representation ``for free'', and there may be less need to posit that it’s part of core knowledge.

\section{General Discussion}

%%% Boccato et al., 2021: Boccato, T., Testolin, A., & Zorzi, M. (2021). Learning numerosity representations with transformers: number generation tasks and out-of-distribution generalization. Entropy, 23, 857.
%%% Kim et al., 2021: Kim, G., Jang, J., Baek, S., Song, M., & Paik, S. B. (2021). Visual number sense in untrained deep neural networks. Science Advances, 7, eabd6127.
Prior studies have used computer vision models to investigate mathematical thinking~\cite{Boccato2021,Kim2021,Nasr2019-xl,Stoianov2012,Testolin2020,upadhyay2023cnn,Upadhyay2025,wang2025computervisionmodelshumanlike,Zorzi2017-wx}. Most have focused on adult cognition, with only \citet{Testolin2020} exploring the question of cognitive development. However, this study suffers from several limitations: It utilized a custom deep neural network as opposed to a standard CNN or vision transformer architecture, it used a custom ‘layer-wise’ training procedure rather than a standard procedure, and it used a custom training dataset of abstracted stimuli rather than naturalistic images. The current study takes an important step beyond this earlier work.

We asked whether a standard CV model architecture trained on a standard image dataset shows human-like trajectories in the growth of GT concept sensitivity and number representation precision. We chose the ResNet-50 model for this case study because prior research has demonstrated its cognitive alignment with how adults represent geometric concepts \cite{Upadhyay2025}. We trained it on the ImageNet image dataset~\cite{imagenet} and saved model checkpoints along the way. Experiment 1 found increasing sensitivity to four classes of GT concepts over training -- Euclidean Geometry, Geometric Figures, Metric Properties, and Topology -- mimicking the trajectory observed in humans (albeit with lower overall performance). However, there was no improvement with training for the three other classes -- Chiral Figures, Geometric Transformations, and Symmetrical Figures. Experiment 2 probed the development of number representations. That humans understand numbers by reference to an MNL is evidenced by the distance, size, and ratio effects. Moreover, these effects sharpen over development, signaling an increase in the precision of this representation. This was also the case for ResNet-50 over the course of training, most strongly for the distance and ratio effects and for stimulus sets 1 -- 4.

An important question for developmental science is: \emph{what develops?} Experiment 1 gives suggestive but no definitive answer for GT concepts: some GT concepts might come ``for free'' from learning to perceive the world, whereas other concepts appear not to be so easily learnable. This might signal that these latter concepts are part of the child’s core knowledge’ \cite{Spelke2007}, and thus the mind/brain does not have to be architected to learn them from experience. (Another interpretation is that they do not belong to core knowledge either, and instead must be learned from supervised mathematics instructions.) Experiment 2 gives a clearer answer to the question: what develops is the model’s latent number line representation, which becomes increasingly canonical over training; see Figure \ref{fig:mil}. This is the same ``mechanism of change'' proposed by mathematical development researchers
~\cite{Halberda2008-qa,Sekuler1977,Moore2015-yq}.

Together, these results show the continuing promise of computer vision models for advancing developmental science. However, for this potential to be realized, several limitations must overcome.

The first limitation concerns the assumption that training on the image classification task is a valid proxy for humans learning to perceive the visual world. This is almost certainly not the case. Vision is useful for object recognition, to be sure, but also for many other functions, such as tracking the movement of objects in space (visual attention)~\cite{Corbetta2002,Szczepanski2013} and reasoning about visuospatial problems (e.g., mental rotation)~\cite{Zacks2008,Tomasino2016}. This gap presents an opportunity. The current study failed to find evidence of growing sensitivity to three classes of GT concepts over training. Perhaps this failure reflects the limits of the image classification task. Future work should explore training CV models on a range of tasks more representative of the range of tasks the human visual system can perform. It may be that additional classes of GT concepts are learned ``for free'' under such an expanded training procedure.

%%% XXX dataset (CITE): There are such datasets, I believe. My memory is that Microsoft has a "things" database with the number of objects in each image coded. I also think the Testolin ... McClelland (2020) study used an image database where each object was surrounded by a bounded box so that the number of objects in each image was known. Let's cite some of these here.
A second limitation is the limited nature of the mathematical measures used. In testing sensitivity to GT concepts, each concept was represented by only one stimulus. It is possible that the 5 images that embody a concept also shared other perceptual properties which are not present in the odd-one-out image, and that these properties instead drove model performance. A stronger benchmark would include many more stimuli for each concept. Experiment 2 used a broader range of stimuli (6 sets) to evaluate the development of number representations over training. The distance and ratio effects were weakest for the most varied stimulus sets (5 and 6) that used the most ``naturalistic'' stimuli, including clip art images from a Google Image search. Future work should use even more visually complex images, such as stimuli from the MS COCO~\cite{lin2015microsoftcococommonobjects}
% Visual Genome~\cite{krishna2016visualgenomeconnectinglanguage}, 
and CLEVR~\cite{johnson2016clevrdiagnosticdatasetcompositional} datasets, to further test the robustness and generalization of the latent number representation learned by CV models. 

A third limitation is that this case study explored only one model architecture, a CNN trained on one image dataset. It likely underestimates the potential developmental alignment of CV models. Future research should explore a variety of model architectures trained on a range of datasets. For example, \citet{wang2025computervisionmodelshumanlike} found that vision transformers like ViT and DINOv2 achieve higher overall accuracy than CNNs on the odd-one-out task for GT concepts, closer to that of young children. At the finer grain of the 7 classes, the correlations between these models and the young children were exceptionally high ($r > 0.90$). It is possible that over training, vision transformer models may show strong alignment to the developmental trajectories of children.

\bibliography{aaai2026}


% \clearpage
\appendix
% \section{Additional information on GT concept human data}

%%% SV: Do we need the "Group Number" column? Relatedly, I think the caption shoul be "The 2 training concepts, the 43 test concepts, and the 7 classes to which the test concepts belong."
\clearpage
% \input{sup}
% \documentclass[letterpaper]{article} % DO NOT CHANGE THIS
% \usepackage{algorithm}
% \usepackage{algorithmic}

% \usepackage{booktabs}       % professional-quality tables
% \usepackage{amsfonts}       % blackboard math symbols
% \usepackage{nicefrac}       % compact symbols for 1/2, etc.
% \usepackage{microtype}      % microtypography
% \usepackage{xcolor}         % colors

% \usepackage{rotating}
% \usepackage{tcolorbox}
% \usepackage{algorithm}
% \usepackage{algpseudocode}
% \usepackage{amsmath}
% \usepackage{algorithmicx}
% \usepackage{booktabs}
% \usepackage{multirow}
% % \usepackage{multicol}
% \usepackage{amsfonts}
% \usepackage{cleveref}
% \begin{document}

\section{GT concepts and their classs for Exp. 1}
\begin{table}[!ht]
\centering
\scriptsize
% \small
\renewcommand{\arraystretch}{1.2}
\begin{tabular}{ll}
\toprule
\textbf{Concept Category} & \textbf{Concept Label} \\
\midrule
(training)                  & Color                                 \\
(training)                  & Orientation                           \\
Topology                    & Holes                                 \\
Topology                    & Inside                                \\
Topology                    & Closure                               \\
Topology                    & Connectedness                         \\
Euclidean geometry          & Alignment of points in lines         \\
Euclidean geometry          & Curve                                 \\
Geometrical figures         & Convex shape                          \\
Euclidean geometry          & Straight line                         \\
Euclidean geometry          & Alignment of points in lines         \\
Geometrical figures         & Quadilateral                          \\
Geometrical figures         & Right angled triangle                \\
Euclidean geometry          & Right angle                           \\
Euclidean geometry          & Right angle                           \\
Metric properties           & Distance                              \\
Geometrical figures         & Circle                                \\
Metric properties           & Center of circle                      \\
Metric properties           & Middle of segment                     \\
Geometrical figures         & Equilateral triangle                  \\
Metric properties           & Fixed proportion                      \\
Metric properties           & Center of quadilateral                \\
Geometrical figures         & Square                                \\
Geometrical figures         & Rectangle                             \\
Geometrical figures         & Parallelogram                         \\
Geometrical figures         & Trapezoid                             \\
Geometrical transformations & Vertical symmetry                     \\
Symmetrical figures         & Vertical axis                         \\
Symmetrical figures         & Horizontal axis                       \\
Symmetrical figures         & Oblique axis                          \\
Geometrical transformations & Translation                           \\
Geometrical transformations & Point symmetry                        \\
Geometrical transformations & Horizontal symmetry                   \\
Geometrical transformations & Rotation                              \\
Geometrical transformations & Oblique symmetry                      \\
Geometrical transformations & Homothecy (fixed orientation)         \\
Euclidean geometry          & Parallel lines                        \\
Chiral figures              & Oblique axis                          \\
Geometrical transformations & Homothecy (fixed size)                \\
Euclidean geometry          & Secant lines                          \\
Chiral figures              & Vertical axis                         \\
Chiral figures              & Vertical axis                         \\
Metric properties           & Equidistance                          \\
Chiral figures              & Oblique axis                          \\
Metric properties           & Increasing distance                   \\
\bottomrule
\end{tabular}
\caption{The 2 training concepts, the 43 test concepts, and the 7 categories to which the test concepts belong.}
\label{tab:gt-concepts}
\end{table}
% \clearpage

\section{Histogram of participant ages for Exp. 1}
\begin{figure}[!ht]
    \centering
    \includegraphics[width=1\linewidth]{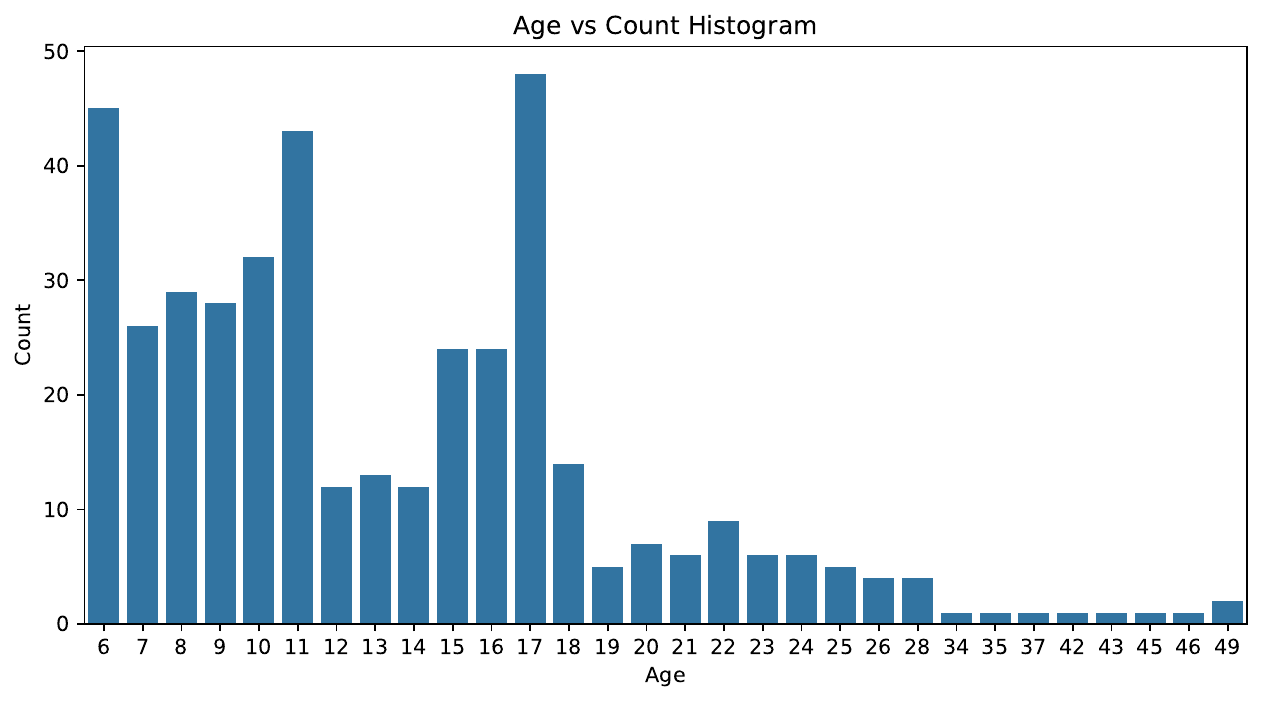}
    \caption{Histogram of the ages of participants in Experiment 2 of \citet{Izard2009-em}.}
    \label{fig:gt-hist}
\end{figure}

\section{Growth function results for Exp. 1}
\begin{table}[ht]
\centering
\scriptsize
\renewcommand{\arraystretch}{1.2}
\begin{tabular}{lccc}
\toprule
\textbf{Category} & \(r\) (p) & \(R^2_{\text{Izard}}\) & \(R^2_{\text{Model}}\) \\
\midrule
Chiral figures             & \(-0.0124\) (0.9481)     & 0.3262 & 0.0088 \\
Euclidean geometry         & \(0.5049\) (0.0044) & 0.4054 & 0.6168 \\
Geometrical figures        & \(0.5434\) (0.0019) & 0.4191 & 0.3928 \\
Geometrical transformations & \(-0.1936\) (0.3053)     & 0.0942 & 0.0486 \\
Metric properties          & \(0.2255\) (0.2308)      & 0.3819 & 0.2226 \\
Symmetrical figures        & \(-0.0285\) (0.8812)     & 0.7043 & 0.0008 \\
Topology                   & \(-0.0482\) (0.8003)     & 0.0765 & 0.1301 \\
\bottomrule
\end{tabular}
\caption{Statistical comparison between human performance from \citet{Izard2009-em} and model performance by class. $r$ ($p$): Pearson $r$ and $p$ value between human data and model data across age / epoch. \(R^2_{\text{Izard}}\): fit of a power function to the human development data. \(R^2_{\text{Model}}\): fit of a power function to the model training data.}
\label{tab:gt-addi-results}
\end{table}

\section{Growth function results for Exp. 2}

\begin{table}[ht]
\centering
\small
\begin{tabular}{lccc}
\toprule
\textbf{Stimulus Set} & \(R^2_{\text{Dist}}\) & \(R^2_{\text{Size}}\) & \(R^2_{\text{Ratio}}\) \\
\midrule
Stimulus Set 1 & 0.49 & 0.04 & 0.35 \\
Stimulus Set 2 & 0.43 & 0.00 & 0.39 \\
Stimulus Set 3 & 0.39 & 0.13 & 0.36 \\
Stimulus Set 4 & 0.38 & 0.25 & 0.48 \\
Stimulus Set 5 & 0.00 & 0.26 & 0.15 \\
Stimulus Set 6 & 0.12 & 0.15 & 0.01 \\
\bottomrule
\end{tabular}
\caption{$R^2$ of power function fit for each stimulus set on Distance, Size, Ratio effects over developmental trajectories.}
\label{tab:num-addi-results}
\end{table}

% \end{document}

% \section{Examples of numerosity stimulus}
% \input{ReproducibilityChecklist/LaTeX/ReproducibilityChecklist}
\end{document}